# A Computational Approach for Human-like Motion Generation in Upper Limb Exoskeletons Supporting Scapulohumeral Rhythms

Rana Soltani-Zarrin, Amin Zeiaee, Reza Langari, Reza Tafreshi

*Abstract*—This paper proposes a computational approach for generation of reference path for upper-limb exoskeletons considering the scapulohumeral rhythms of the shoulder. The proposed method can be used in upper-limb exoskeletons with 3 Degrees of Freedom (DoF) in shoulder and 1 DoF in elbow, which are capable of supporting shoulder girdle. The developed computational method is based on Central Nervous System's (CNS) governing rules. Existing computational reference generation methods are based on the assumption of fixed shoulder center during motions. This assumption can be considered valid for reaching movements with limited range of motion (RoM). However, most upper limb motions such as Activities of Daily Living (ADL) include large scale inward and outward reaching motions, during which the center of shoulder joint moves significantly. The proposed method generates the reference motion based on a simple model of human arm and a transformation can be used to map the developed motion for other exoskeleton with different kinematics. Comparison of the model outputs with experimental results of healthy subjects performing ADL, show that the proposed model is able to reproduce human-like motions.

## I. INTRODUCTION

Exoskeletons' ability to provide high intensity and long therapy sessions, accurate measurements and precise control of individual joints, make them a great adjunct to manual rehabilitation. An important requirement for using exoskeleton based systems is development of a reference behavior for the device. Some of the current methods necessitate the presence of a therapist to move the patient's arm to record *every* single motion [1], and/or require bilateral exoskeletons or a motion capture system. Computational methods can facilitate the path generation process by eliminating the need for a priori recording of each motion. Existing computational methods for reference generation, are derived based on the fixed shoulder assumption [2] and are verified experimentally for simple, low RoM reaching motions where the contribution of the scapulohumeral rhythms is negligible. However, the assumption of fixed shoulder is not valid in upper-limb motions involving high arm elevations such as Activities of Daily Living (ADL) [3].

Neglecting scapulohumeral rhythms in generation of the reference trajectory for a rehabilitation exoskeleton has undesirable consequences such as misalignment between the joint axis/center of the exoskeleton and anatomical joint of the human body. Joint misalignment can cause discomfort to the patient due to residual forces at human robot interaction points [4]. This paper presents a computational path generation approach for exoskeletons supporting inner shoulder motion [5, 6], and is capable of producing human like motions. Considering the role of inner shoulder in ADL motions, and the therapeutic value of practicing ADL tasks in functional recovery of stroke patients, the proposed method can be deployed in rehabilitation exoskeletons.

## II. PROPOSED METHOD

Most ADL motions require large range of movement and involve high arm elevation. Elevation of the arm yields movement of the center of the Glenohumeral (GH) joint through the scapulohumeral rhythms, as shown in Fig.1.

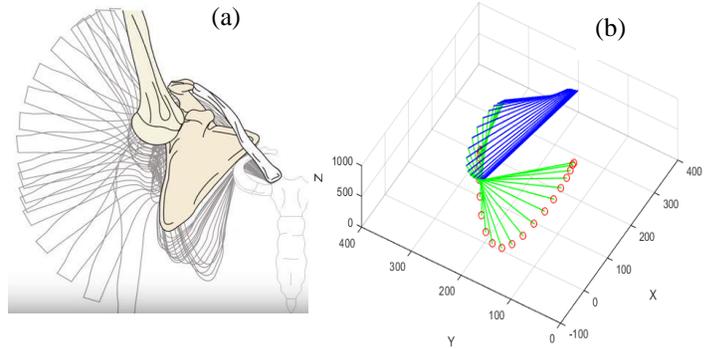

Figure 1. (a) GH motion in arm elevation [Frederic Rf, fisiohipotesis], (b) experimental data of combing motion captured by Vicon system [8]

To include GH joint center movement in the analysis, an analytic model [4, 9] is adopted, which provides the position of the GH joint center in spherical coordinates using the length of the vector connecting the fixed origin to the GH joint center, $d_{SG}$, and the angles of elevation/depression, $\varphi_{ed}$, and protraction/retraction inclinations of this vector, $\varphi_{pr}$, as shown in Fig. 2. To preserve smoothness in the analysis, a polynomial fit of the model is derived as follows:

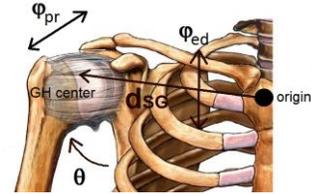

Figure 2. Upper limb model and angles

$$\begin{cases} \varphi_{ed} = 1.49 \times 10^{-9} \theta^5 - 4.28 \times 10^{-7} \theta^4 + 1.44 \times 10^{-5} \theta^3 + 5.2 \times 10^{-3} \theta^2 \\ \quad\quad -0.1357\theta + 0.7078 \\ \varphi_{pr} = 1.82\theta^3 - 8.073\theta^2 - 3.99\theta \end{cases} \quad (1)$$

*Research supported by Qatar National Research Fund (7–1685–2–626).

R. Soltani-Zarrin, A. Zeiaee, R. Langari are with the Mechanical Engineering Department, Texas A&M University, College Station, TX (e-mail: rana.soltani@tamu.edu; amin.zeiaee@tamu.edu; rlangari@tamu.edu ).
R. Tafreshi is with the Department of Mechanical Engineering, Texas A&M University at Qatar, (e-mail: reza.tafreshi@qatar.tamu.edu).

$$d_{SG} = d_0(-1.6 \times 10^{-5}\theta^2 + 3 \times 10^{-4}\theta + 1)$$

where $\theta$ is the elevation of the arm, and $d_0 = d_{SG}|_{\theta=0}$

Given the initial and final positions of the hand, to construct a motion for the exoskeleton close to the natural motion of the human arm, this paper adopts geodesic curves in the Riemannian space for path generation in the configuration space [2,10]. Using the mass-inertia matrix of arm, $M$, as the tensor of the Riemannian space, and solving the following dynamical equation subject to boundary conditions, the geodesic curves can be found:

$$M(q)q'' + C(q,q')q' = 0 \qquad q(0) = q_0, q(1) = q(\alpha, x_f) \qquad (2)$$

where the prime sign denotes the derivative with respect to the path parametrization variable $\lambda$, and $C(q,q')$ represents the coriolis matrix. As equation (2) shows to solve the geodesic equation, the final configuration of the arm is needed. Determining the final configuration requires the solution of the inverse kinematics [8,9] which depends on the position of the shoulder center. Denoting the positions of the GH joint center, elbow and wrist as $X_{sh}(\theta)$, $X_e$, and $X_w$, respectively, and using the model in (1), (3) shows inverse kinematics formulations for hand, where $q = (\theta, \eta, \zeta, \phi)$ denotes the arm configuration, and $l_u$, $l_f$ are upper-arm, and forearm lengths.

$$\theta = \cos^{-1}\left(\frac{z_{sh}(\theta) - z_e}{l_u}\right), \quad \eta = atan2(x_{sh}(\theta) - x_e, y_e - y_{sh}(\theta))$$
$$\zeta = atan2\big(l_u((x_e - x_{sh})(y_w - y_{sh}) - (x_w - x_{sh})(y_e - y_{sh})), (y_e - y_{sh})((y_e - y_{sh})(z_w - z_{sh}) - (y_w - y_{sh})(z_e - z_{sh})) - (x_e - x_{sh})((z_e - z_{sh})(x_w - x_{sh}) - (z_w - z_{sh})(x_e - x_{sh}))\big) \qquad (3)$$
$$\phi = \cos^{-1}\left(\frac{(x_w - x_{sh})^2 + (y_w - y_{sh})^2 + (z_w - z_{sh})^2 - l_u^2 - l_f^2}{2l_u l_f}\right)$$

Solving the geodesic equations for each value of the final configuration that satisfies the physiological limits on the arm joint, defines a family of paths in the configuration space parameterized by the swivel angle, $\alpha$. Choice of $\alpha$, or alternatively the final configuration of the arm, is based on the minimum energy concept.

The proposed method uses a model of human arm parametrized in arm elevation, shoulder internal/external rotation and shoulder horizontal abduction/adduction [10]. Based on the geometric relations between the angles of the consecutive axes in the shoulder structure of any exoskeleton, a transformation can be defined relating the arm model used here and the kinematics of the exoskeleton shoulder. Using this transformation, $T: q_h \rightarrow q_r$, developed path can be generated in the exoskeleton joint space.

### III. SIMULATION AND EXPERIMENTAL RESULTS

To verify accuracy of the developed computational path generation method with respect to natural human arm motion, 6 healthy subjects were recruited and 4 types of ADL motions, with total 120 trials, were captured by a Vicon motion capture system [9]. Fig. 3 shows the comparison between experimental results and outputs of the computational model for a subject performing "reaching high". The quantitative analysis for all 120 trials, show an average of 0.92 for $r^2$ measure [9].

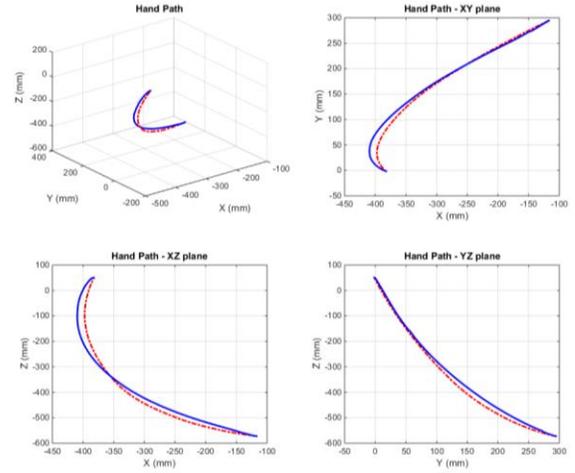

Figure 3. Riemmanian geodesic for reaching high

### IV. CONCLUSION

In this paper a new computational method is introduced for generating reference paths for ADL motions in upper-limb exoskeletons that support inner shoulder motion. To the best of our knowledge this is the first exoskeleton path generation method for ADL tasks that considers scapulohumeral rhythms. Accuracy of the proposed model is examined by comparing the simulation results with the experimental data collected form healthy subjects performing ADL tasks.